\documentclass[sigconf]{acmart}
\AtBeginDocument{%
  }

\renewcommand\footnotetextcopyrightpermission[1]{} 
\acmConference[ACM e-Energy 2026]{The 17th ACM International Conference on Future and Sustainable Energy Systems}{June 2026}{Banff, Canada}
\acmBooktitle{Proceedings of the 17th ACM International Conference on Future and Sustainable Energy Systems (e-Energy ’26), June 22--25, 2026, Banff, Canada}

\usepackage{booktabs} 
\usepackage{multirow}
\usepackage[utf8]{inputenc}
\usepackage{graphicx}
\usepackage{hyperref}
\usepackage{caption}
\usepackage{amsthm}
\usepackage{tabularx} 
\usepackage{enumitem}
\theoremstyle{remark}
\usepackage{xcolor}
\newtheorem*{remark}{Remark}
\usepackage[most]{tcolorbox}
\usepackage{subcaption} 

\newcommand{\nop}[1]{}
\newcommand{\sysname}{\textsf{CaberNet}}

\begin{document}
\title{\sysname{}: Causal Representation Learning for Cross-Domain HVAC Energy Prediction}

\author{Kaiyuan Zhai}
\email{rickzky1001@gmail.com}
\affiliation{
  \institution{The Hong Kong University of Science and Technology (Guangzhou)}
  \city{Guangzhou}
  \country{China}}
  
\author{Jiacheng Cui}
\email{jiachengc648@gmail.com}
\affiliation{
  \institution{The Hong Kong University of Science and Technology (Guangzhou)}
  \city{Guangzhou}
  \country{China}}
  
\author{Zhehao Zhang}
\email{zhang.zheha@northeastern.edu}
\affiliation{
  \institution{The Hong Kong University of Science and Technology (Guangzhou)}
  \city{Guangzhou}
  \country{China}}
  
\author{Junyu Xue}
\email{junyuxue@outlook.com}
\affiliation{
  \institution{Southern University of Science and Technology}
  \city{Shenzhen}
  \country{China}}
  
\author{Yang Deng}
\email{yang2.deng@connect.polyu.hk}
\affiliation{
  \institution{The Hong Kong Polytechnic University}
  \country{Hong Kong}}

\author{Kui Wu}
\email{wkui@uvic.ca}
\affiliation{
  \institution{University of Victoria}
  \city{Victoria}
  \country{Canada}}

\author{Guoming Tang}
\authornote{corresponding author}
\affiliation{
  \institution{The Hong Kong University of Science and Technology (Guangzhou)}
  \city{Guangzhou}
  \country{China}}
\email{guomingtang@hkust-gz.edu.cn}

\renewcommand{\shortauthors}{Zhai et al.}

\begin{abstract}
Cross-domain HVAC energy prediction is essential for scalable building energy management, particularly because collecting extensive labeled data for every new building is both costly and impractical. Yet, this task remains highly challenging due to the scarcity and heterogeneity of data across different buildings, climate zones, and seasonal patterns. In particular, buildings situated in distinct climatic regions introduce variability that often leads existing methods to overfit to spurious correlations, rely heavily on expert intervention, or compromise on data diversity. To address these limitations, we propose \sysname{}, a causal and interpretable deep sequence model that learns invariant (Markov blanket) representations for robust cross-domain prediction. In a purely data-driven fashion and without requiring any prior knowledge, \sysname{} integrates i) a global feature gate trained with a self-supervised Bernoulli regularization to distinguish superior causal features from inferior ones, and ii) a domain-wise training scheme that balances domain contributions, minimizes cross-domain loss variance, and promotes latent factor independence. We evaluate \sysname{} on real-world datasets collected from three buildings located in three climatically diverse cities, and it consistently outperforms all baselines, achieving a 22.9\% reduction in normalized mean squared error (NMSE) compared to the best benchmark. Our code is available at \url{https://github.com/SusCom-Lab/CaberNet-CRL}.

\end{abstract}




\maketitle

\section{Introduction}
Heating, Ventilation, and Air Conditioning (HVAC) systems are widely used in residential, commercial, and industrial buildings, accounting for a substantial portion of total energy consumption~\cite{zhang2025climate}. In office buildings, this figure can reach approximately 40\%~\cite{australiaHVAC}. Accurate prediction of HVAC energy usage is therefore critical for enhancing automated control systems~\cite{chennapragada2022time} and advancing sustainable building management practices~\cite{pinheiro2025ahp}. Nowadays, with the proliferation of smart meters and IoT sensors, large volumes of building-level data are readily available, offering significant opportunities for data-driven predictive modeling. 

Recent approaches have shown promising results in modeling HVAC systems under the independent and identically distributed (i.i.d.) assumption, using methods ranging from traditional machine learning~\cite{cui2024energy, mohan2025ensemble} and deep learning~\cite{ni2024study,ciampi2024energy, asamoah2025evaluating,liu2024energy} to
physics-informed methods~\cite{ma2024physics,chen2023physics} and transfer learning~\cite{hooshmand2019energy,raisch2025gentl, xing2024transfer}.
However, a fundamental limitation of most of these methods is their reliance on Empirical Risk Minimization (ERM), which learns purely from statistical correlations and often captures domain-specific patterns rather than underlying \emph{causal mechanisms}\footnote{The \textit{causal mechanism} refers to a stable process in which causes give rise to effects. Mathematically, this is often represented as $P(\text{effect} \mid \text{cause}).$}. For instance, a model may learn a temperature-load correlation that holds during summer but fails to generalize to winter conditions. Such spurious associations deviate from the true causal drivers of energy consumption, leading to poor transferability to new domains with different spatial or temporal contexts. Consequently, achieving robust generalization with these approaches would require training on abundant data from diverse domains~\cite{scholkopf2021toward}, while transfer learning also requires a non-trivial amount of target-domain data for fine-tuning~\cite{hooshmand2019energy,raisch2025gentl}. These dependencies significantly limit the feasibility and practicality of deploying existing methods in real-world scenarios.

This challenge motivates the development of models that can be trained on buildings with available data and applied directly to others, a setting known as \emph{domain generalization (DG)}. In practice, however, this is difficult due to the high diversity inherent in buildings and their operating conditions~\cite{hu2021building}. Some sources of variation are directly observable and can be measured in datasets, such as local climate, interior layout, and building materials. By contrast, many others are latent and hard to record, like occupant behavior and maintenance quality, and thus often absent from datasets, complicating the learning of robust, domain-invariant representations. 

Causal Machine Learning (CML) has recently demonstrated strong potential for tackling out-of-distribution (OOD) challenges in domain generalization, with successful applications in areas such as computer vision~\cite{nguyen2023causal,lv2022causality,liu2024causality} and HVAC system modeling~\cite{jiang2025if,duhirwe2024causal,huang2024entropy}. The core premise of CML is that: \textit{robust generalization under distribution shift requires uncovering the latent mechanisms that generate the data, rather than fitting to surface-level correlations~\cite{scholkopf2021toward}}. In this view, predictors should depend on invariant causes ($X_{S^*}$) of the target variable ($Y$), ensuring that the conditional distribution $P(Y\,|\,X_{S^*})$ remains stable across environments. Accordingly, the CML pipeline aims to i) identify and extract features that capture stable, mechanism-level structures shared across domains, consistent with the principle of Invariant Risk Minimization (IRM)~\cite{arjovsky2019invariant}, and ii) attenuate the influence of domain-variant spurious features~\cite{ma2024causality}.

Despite this promise, existing CML methods still face practical limitations. A common line of work relies on prior-knowledge-based causal interventions, e.g., by applying the do-operator to image styles or backgrounds that are known to be unrelated to the classification label~\cite{lv2022causality,liu2024causality}. Other approaches embed more specialized expert knowledge into the model~\cite{huang2024entropy,duhirwe2024causal}. Yet such strategies are infeasible in domains where prior knowledge is limited or uncertain, and also undermine generalizability to new problem settings. In the context of HVAC systems, for instance, even basic questions, such as \emph{how} and \emph{to what extent} occupant behaviour affects HVAC energy consumption, are difficult to specify a priori. A second line of work attempts to sidestep this issue through data filtering, selectively training on samples from similar distributions~\cite{jiang2025if,xing2024transfer}. While this can reduce the domain gap, it also sacrifices distributional diversity. This loss of diversity is counterproductive for domain generalization, as contrasting environments are essential for discovering causal features. Furthermore, such filtering is particularly problematic when data are already scarce, as it further diminishes the limited information available for learning.

In this work, we therefore revisit the challenge of cross-domain HVAC energy prediction through the lens of causal representation learning (CRL). We propose \sysname{} (\textbf{\underline{Ca}}usal \textbf{\underline{Ber}}noulli \textbf{\underline{Net}}work), a novel framework that learns causal representations directly from raw data without requiring prior knowledge or discarding valuable samples. Concretely, our approach integrates three key innovations, where we i) introduce a global feature gate that assigns consistent importance weights across domains, thereby providing interpretability of which raw features drive energy consumption, ii) develop a pure data-driven self-supervised Bernoulli regularization that softly partitions features into higher and lower importance groups, without the need for prior knowledge or labels, and iii) design domain-wise training objectives that reweight per-domain difficulty, penalize cross-domain loss variance, and encourage independence of latent factors.
In combination with the task loss, \sysname{} guides learning toward invariant, mechanism-level relations rather than domain-specific shortcuts or correlations, enabling robust out-of-domain generalization.

Our contributions are summarized as follows:
\begin{itemize}[leftmargin=*]
    \item \emph{A causal representation learning framework.} We introduce a unified framework that integrates causal principles with deep sequence modeling, enabling the disentanglement of raw features into Markov blanket causal representations for robust cross-domain prediction under distribution shifts.
    \item \emph{Self-supervised feature selection.} We propose a novel regularization method that combines $\ell_1$ sparsity with Bernoulli-entropy minimization to automatically identify stable causal features without supervised labels.
    \item \emph{Invariant domain-wise training.} We design a training strategy that aggregates per-domain losses with difficulty weights, penalizes cross-domain loss dispersion, and promotes latent factor independence.
    \item \emph{Empirical validation and explainability.} We demonstrate that \sysname{} achieves a 22.9\% reduction in NMSE over the state-of-the-art benchmarks on real-world data from six office floors across three cities, while offering high explainability consistent with HVAC domain knowledge.
\end{itemize}

\section{Causal Perspective on Energy Prediction}
\subsection{Rationale}
\label{sec:rationale}

Causal inference aims to uncover and model the cause-effect relationships among variables, enabling predictions that remain valid under distribution shifts~\cite{pearl2009causality}. Unlike purely statistical methods that rely on correlations, causal approaches seek to identify stable relationships that reflect the underlying data-generating process, thereby improving generalization across unseen domains.

This distinction is critical for energy prediction in HVAC systems. Consider the relationship between outdoor temperature and energy consumption: a spike in temperature during summer increases cooling demand, raising energy usage; a drop in temperature during winter increases heating demand, also raising energy usage. While a correlation exists, the effect of temperature on HVAC energy demand is domain-variant. A model that learns only this seasonal correlation may fail to generalize, as it learns a domain-specific shortcut rather than the underlying causal mechanism. A robust model must instead identify the invariant causes whose relationship with energy consumption remains stable.

Guided by this rationale, we base our approach on three key concepts from causal reasoning.

\paragraph{\textbf{Guideline 1. Causal Invariance Assumption~\cite{peters2016causal,pfister2019invariant}}}
\label{para:cia}
The foundation of our approach is the principle that robust prediction requires identifying a stable causal core. Invariant Causal Prediction (ICP) formalizes this by seeking a subset of predictors $S^* \subseteq \{1, \dots, p\}$ whose conditional relationship with the target variable $Y$ remains stable across multiple environments $e \in \mathcal{E}$. This invariant set $S^*$ (or invariant causal predictors) is crucial for ensuring generalizability under distributional shifts.

This is highly relevant to HVAC systems. While different buildings (domains $e$) exhibit vastly different joint distributions of features $X$ (e.g., due to climate or occupancy differences), the underlying physical and operational mechanisms governing energy consumption are mainly stable. This means the relationship between the true causal drivers $X_{S^*}$ and energy usage $Y$ is invariant:
\[
P\left(Y^e \mid X^e_{S^*}\right) = P\left(Y^{e'} \mid X^{e'}_{S^*}\right), \quad \forall e, e' \in \mathcal{E}
\]
while in general,
\[
P\left(Y^e \mid X^e\right) \neq P\left(Y^{e'} \mid X^{e'}\right), \quad \forall e, e' \in \mathcal{E}
\]
as the full set of predictors $X$ may include domain-variant, spurious factors whose effects differ across environments.

Therefore, our goal is to discover $S^*$ rather than simply rely on all available predictors.

\begin{figure}[t]
  \centering
  \includegraphics[width=0.55\linewidth]{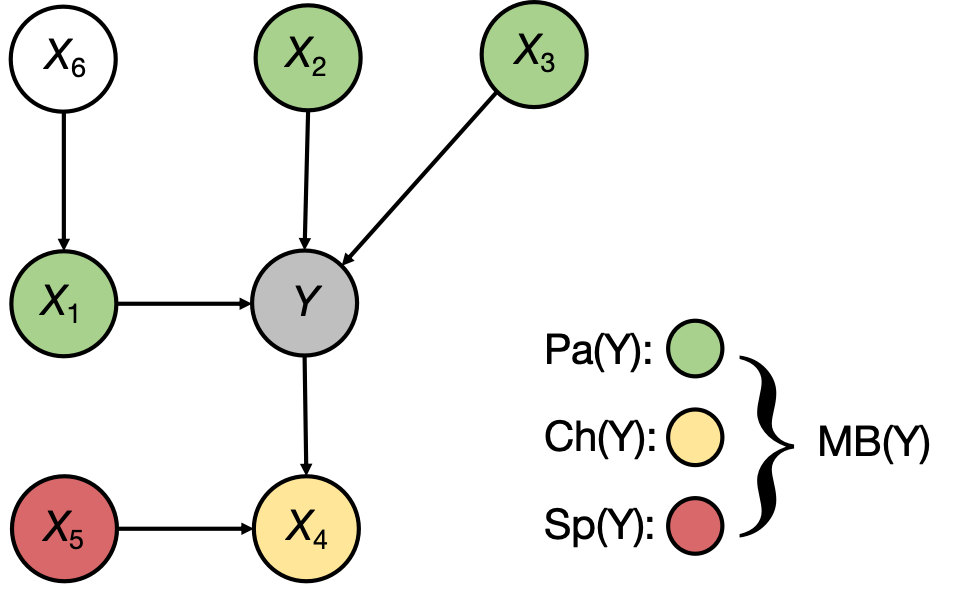}
  \caption{Illustration of Markov blanket (Pa: parent; Ch: child; Sp: spouse).}
  \label{fig:mkb}
  \vspace{-10pt}
\end{figure}

\paragraph{\textbf{Guideline 2. Markov Blanket Prediction~\cite{jiang2024improving,yin2024integrating}}}
\label{para:g2mb}
The Markov blanket (MB) of a target variable $Y$, denoted as $\mathrm{MB}(Y)$, is the minimal set of variables that renders $Y$ conditionally independent of all other variables in the system~\cite{pearl2009causality}. As illustrated by the causal graph in Figure~\ref{fig:mkb}, $\mathrm{MB}(Y)$ consists of: i) the parents of $Y$ (its direct causes, $X_1, X_2, X_3$), ii) the children of $Y$ (its direct effects, $X_4$), and iii) the spouses of $Y$ (other direct causes of $Y$’s children, $X_5$).

Formally, for any variable $O$ outside the blanket and distinct from $Y$ ($X_6$ in Figure~\ref{fig:mkb}):
\begin{equation}
\label{Markov_blanket_definition}
Y \perp\!\!\!\perp O \mid \mathrm{MB}(Y),
\end{equation}
where $\perp\!\!\!\perp$ means ``is independent of''. This guideline means that $\mathrm{MB}(Y)$ contains all necessary information to predict $Y$ and blocks all influence from extraneous variables.

While using only the parents of $Y$ captures the true causal mechanism $P(Y \mid \mathrm{Pa}(Y))$, it may discard valuable predictive signals from its children and their other causes (spouses). For instance, while carrying umbrellas does not cause rain, observing umbrellas can improve the prediction of rain. Therefore, conditioning on the full $\mathrm{MB}(Y)$ could reduce the conditional variance of $Y$:
\begin{equation}
\mathrm{Var}\!\left(Y \mid \mathrm{MB}(Y)\right) \le \mathrm{Var}\!\left(Y \mid \mathrm{Pa}(Y)\right),
\label{eq:Markov_blanket_variance}
\end{equation}
with equality holding only if the children and spouses provide no additional information about $Y$ (see proof in Appendix~\ref{app:lower_variance}).

Conditioning on all variables can be effective in i.i.d. settings.
However, in the presence of domain shifts, a critical distinction emerges: while the $\mathrm{MB}(Y)$ is the \emph{minimal sufficient set} for prediction, the full set of variables $X$ may contain spurious correlations. Thus, in an i.i.d. setting, $\mathrm{Var}\!\left(Y \mid X\right) \leq \mathrm{Var}\!\left(Y \mid \mathrm{MB}(Y)\right)$ may hold due to these additional correlations. Yet, under a distribution shift, these correlations can become unstable and lead to severe performance degradation. Therefore, the $\mathrm{MB}(Y)$ represents the optimal trade-off, as it harnesses more predictive signal than the parent set alone while avoiding the inclusion of spurious, domain-specific noise that plagues the full feature set. This makes it an ideal basis for building robust, cross-domain predictors.

\paragraph{\textbf{Guideline 3. Independence of Causal Representation Assumption~\cite{lv2022causality}}}
\label{par:indy}
To learn a disentangled and compact latent representation, we adopt the principle that the underlying causal mechanisms are independent. We posit that the learned latent factors $Z=[Z_1, \dots, Z_d]$ are independent in the marginal sense:
\begin{equation}
\label{eq:indy}
    p(Z) \;=\; \prod_{i=1}^{d} p(Z_i)
\end{equation}
and that this representation is causally sufficient for prediction~\cite{pearl2009causality}, meaning it captures all information from $X$ relevant to $Y$, i.e.,
$
Y \;\perp\!\!\!\perp\; X \,\mid\, Z,
$ 
This assumption encourages the elimination of redundant latent dimensions and supports compact models. Our ablation studies (Section~\ref{sec:ablation_indy}) confirm that stable performance is achievable even with a small hidden dimension $d$, validating the efficacy of this approach.

\begin{remark}
     \textbf{Tension with Markov blanket}. Markov blankets include the co-parents/spouses of $Y$'s children; such variables are not, in general, mutually independent (e.g., $X_5$ and $X_4$ are not independent, as shown in Figure~\ref{fig:mkb}). Therefore, enforcing strong factor independence in $Z$ can conflict with a fully learned Markov blanket representation. In practice, we balance this trade-off by treating the independence principle as a regularizing prior rather than a strict requirement. The strength of the independence regularizer $\mathcal{L}_{indy}$ (see Eq.~\ref{eq:L_indy}) is tuned relative to other terms. This allows the model to retain essential predictive information from the Markov blanket's dependent components while still discouraging unnecessary entanglement and promoting a compact latent space.
\end{remark}

\subsection{Problem Formulation}

We consider the problem of cross-domain HVAC energy consumption prediction. For each domain $e \in \mathcal{E}$ (e.g., a distinct building or one of its floors), the data consist of multivariate time series:
\[
\mathcal{D}^e = \{(X^e_{t-w:t-1}, Y^e_t)\}_{t=w}^{n_e}, \quad X^e_t \in \mathbb{R}^p, \; Y^e_t \in \mathbb{R}
\]
where $w$ is the size of the time window, $X^e_t$ denotes the observed features (e.g., weather, indoor conditions, etc.) at time $t$, and $Y^e_t$ is the corresponding HVAC energy consumption.

The fundamental challenge is distribution shift: the marginal distribution of features varies significantly across domains due to differences in climate, occupancy patterns, and building properties, i.e., $P^e(X) \neq P^{e'}(X)$ for $e \neq e'$. Consequently, a predictor trained to minimize empirical risk on a source domain often fails to generalize, as it may leverage these spurious, domain-specific correlations.

Our core assumption, grounded in causal reasoning, is that while the inputs $X$ may change, the underlying \emph{causal mechanism} relating the true drivers $X_{S^*}$ to the target $Y$ remains invariant:
\[
P^e\!\left(Y \mid X^e_{S^*}\right) = P^{e'}\!\left(Y \mid X^{e'}_{S^*}\right), \quad \forall e,e' \in \mathcal{E}
\]
where $X_{S^*} \subseteq X$ denotes the causality-related features that govern $Y$ across environments.

\textbf{Goal.} Our objective is to learn a model comprising: i) a \emph{causal representation extractor} $Z = h_\phi(X)$ (Section~\ref{sec:cre}), and ii) an \emph{energy predictor} $\hat{Y} = m_\theta(Z)$ (Section~\ref{sec:energy_predictor}) trained over source domains $\mathcal{E}_{\mathrm{train}}$, validated over $\mathcal{E}_{\mathrm{val}}$, with the goal of generalizing to unseen target domains $\mathcal{E}_{\mathrm{test}}$ without retraining. 
Formally, the model is trained on source domains $\mathcal{E}_{\mathrm{train}}$ to minimize the validation loss, and subsequently evaluated on the unseen test domains $\mathcal{E}_{\mathrm{test}}$, i.e.,
\[
\min_{\theta,\phi}\;
\mathbb{E}_{e \in \mathcal{E}_{\mathrm{test}}}\!
\left[\ell\big(m_\theta(h_\phi(X^e)), Y^e\big)\right]
\]
where $\ell(\cdot,\cdot)$ is a suitable loss, e.g., normalized mean squared error. 

The key difficulty in addressing the problem lies in ensuring the composed function $m_\theta \circ h_\phi$ captures the invariant causal relation rather than domain-specific associations. To achieve this, we translate the aforementioned three causal guidelines from Section~\ref{sec:rationale} into concrete model design choices and regularization strategies, forcing the learner to ignore spurious correlations and focus on the true causal drivers of energy consumption.

\section{\sysname{} Framework}
\begin{figure}[t]
  \centering
  \includegraphics[width=1\linewidth]{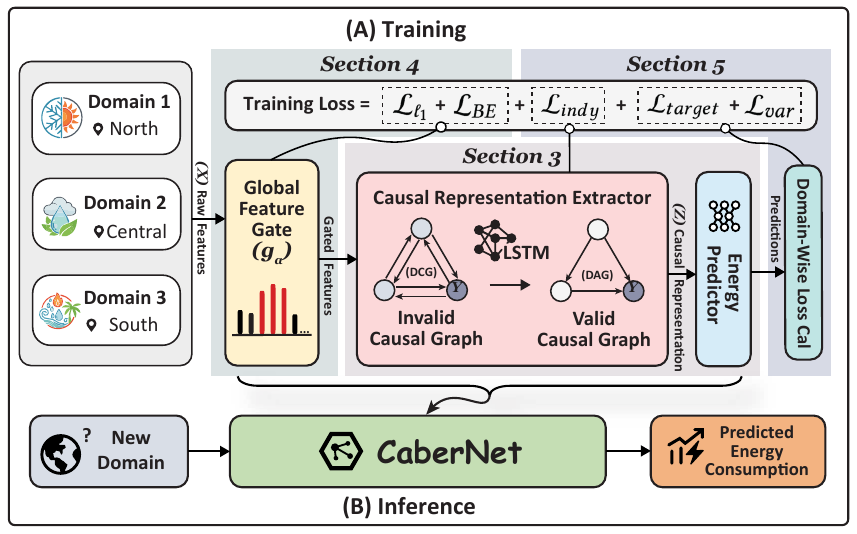}
  \caption{\sysname{} framework. The entire process is end-to-end, and the training loss is computed as a weighted sum of all individual losses. Task: training on the source domain and generalizing to the target domain.}
  \label{fig:framework}
\end{figure}

The overall structure of \sysname{} is shown in Figure~\ref{fig:framework}.
In this section, we provide an overview of this framework by introducing its main modules and their roles.

\subsection{Causal Representation Extractor}
\label{sec:cre}

Structural causal models (SCMs) are typically represented over observable variables within a directed acyclic graph (DAG)~\cite{pearl2009causality}. However, directly applying this framework to HVAC systems is problematic due to the presence of cyclic relationships among observables. For example, indoor temperature and AC energy consumption influence each other in a feedback loop: a higher indoor temperature causes increased AC energy use, which in turn lowers the indoor temperature.

\begin{figure}[t]
  \centering
  \includegraphics[width=0.88\linewidth]{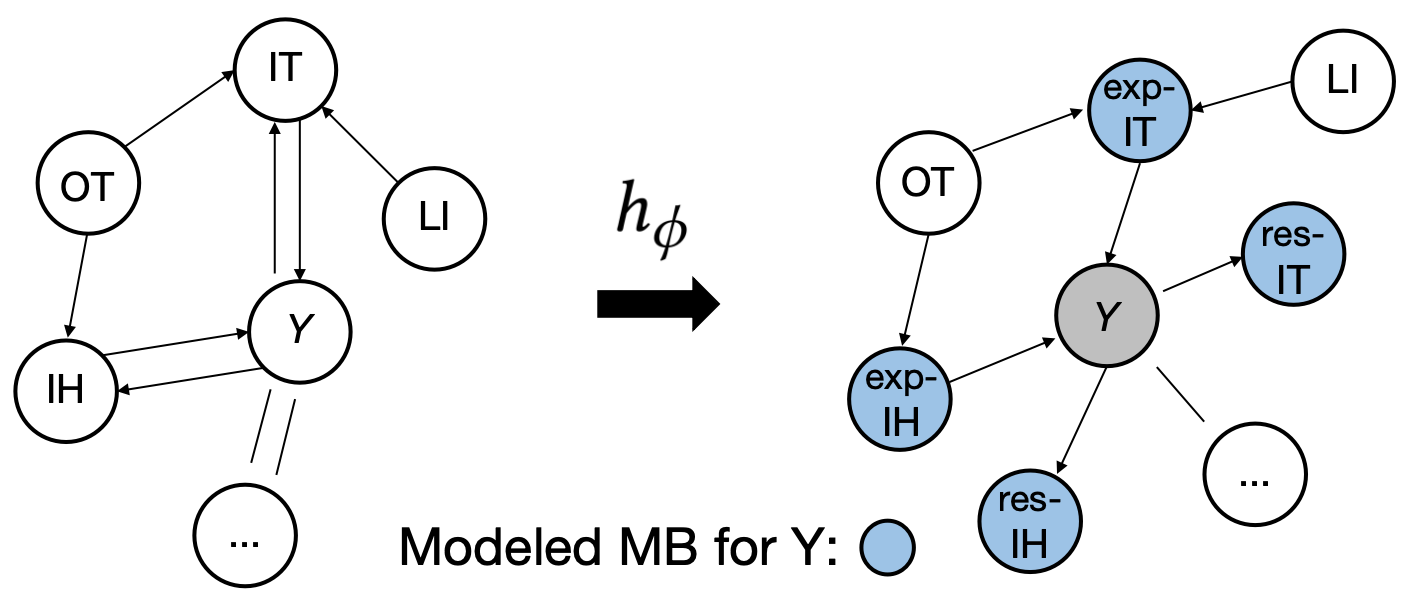}
  \caption{Illustration of reconstructed Markov blanket for $Y$ (IT: indoor temperature, OT: outdoor temperature, IH: indoor humidity, LI: light intensity, exp: explanatory, res: response, MB: Markov blanket). Note that the modelled MB is our target causal representation.}
  \label{fig:modeld_mkb}
\end{figure}

We therefore adopt a decomposition-reconstruction view. Our causal representation extractor $h_\phi$ maps raw, cyclical time-series data into a latent representation $Z$ where the relationships are disentangled and can be represented by a valid SCM (Figure~\ref{fig:framework}). For instance, the single observable ``indoor temperature'' can be decomposed into distinct latent factors: an \emph{explanatory} factor that influences energy usage ($\text{exp-IT} \to Y$ and a \emph{response} factor that is affected by it ($Y \to \text{res-IT}$), as illustrated in Figure~\ref{fig:modeld_mkb}. Such decomposition breaks the cycle and enables causal interpretation.

Our approach diverges from traditional Markov blanket discovery methods~\cite{khan2023novel,li2021Markov}, which perform a combinatorial search-and-prune procedure over raw features. Instead, we implicitly bias the learned representation $Z$ towards the Markov blanket of $Y$ by leveraging its predictive advantage (see Guideline 2 and Eq.~\ref{eq:Markov_blanket_variance}). The reconstructed latent causal graph in Section~\ref{sec:cd} further confirms that the learned representation indeed aligns with a Markov blanket structure. 
In essence, while prior work selects a subset of raw features, our method transforms all features into a latent space where the representation itself aligns with the Markov blanket.

To handle the temporal dependence, we implement $h_\phi$ using a Long Short-Term Memory (LSTM) network~\cite{hochreiter1997long}:
\[
Z \;=\; h_\phi(X_{t-w:t-1}),
\]
where $Z$ is the latent causal representation, and is then passed to the energy predictor $m_\theta$. Note that our framework is \textit{model-agnostic}, and the LSTM here serves as a capable backbone but could be replaced with other sequence models; we adopt LSTM primarily due to the moderate dataset size, which may not support training heavier architectures like Transformers.

\subsection{Energy Predictor}
\label{sec:energy_predictor}

The absolute scale of HVAC energy consumption can vary significantly across domains due to intrinsic factors such as building floor area, HVAC system capacity, and peak occupancy levels. A model that fails to account for these baseline differences may struggle with systematic biases in prediction.

To address this, we explicitly model these domain-specific properties. For each domain, we compute a vector $s$ of descriptive statistics from the training portion of the energy target $Y$:
\[
s \;=\; [\,\mu,\; \sigma,\; q_1,\; q_3\,]^\top \in \mathbb{R}^4,
\]
where $\mu$ is the mean, $\sigma$ the standard deviation, and $q_1$, $q_3$ the first and third quartiles, respectively. This vector $s$ comprehensively captures the central tendency, spread, and distribution shape of energy consumption for a domain.

This vector is then processed by a small Multi-Layer Perceptron (MLP), which acts as a scale encoder, to generate a domain-specific scaling factor $\gamma$ and an intercept $\beta$, i.e.,
\[
(\gamma,\;\beta) \;=\; \text{MLP}(s).
\]

With these parameters, we then apply an affine adjustment to the causal representation $Z$, aligning its scale and offset with the target domain before the final prediction:
\[
\tilde{Z} \;=\; \gamma \cdot Z \;+\; \beta\,,
\qquad
\hat{Y} \;=\;  \mathrm{Linear}(\tilde{Z})
\]

The above design yields two benefits:
\begin{itemize}[leftmargin=*]
    \item \textit{Disentanglement.} It decouples domain-invariant causal mechanisms (learned by $h_\phi$) from domain-specific base energy characteristics, avoiding distortion.
    \item \textit{Generalization.} By normalizing the input to the final predictor across domains, the energy predictor mitigates scale differences, allowing the causal representation to focus on feature-level rather than magnitude-level modeling.
\end{itemize}

\section{Self-Supervised Feature Selection}

A primary challenge in domain generalization is identifying which features encode stable, causal mechanisms versus those that form spurious, domain-specific correlations. To address this without reliance on prior knowledge, we cast feature selection as a self-supervised binary classification problem. The effectiveness of this component is validated by the ablation study contrasting \sysname{} vs. \sysname{}-SIRM in Section~\ref{sec:ablation_reg}.

\subsection{Global Feature Gate}
\begin{figure*}[t]
  \centering
  \includegraphics[width=0.99\linewidth]{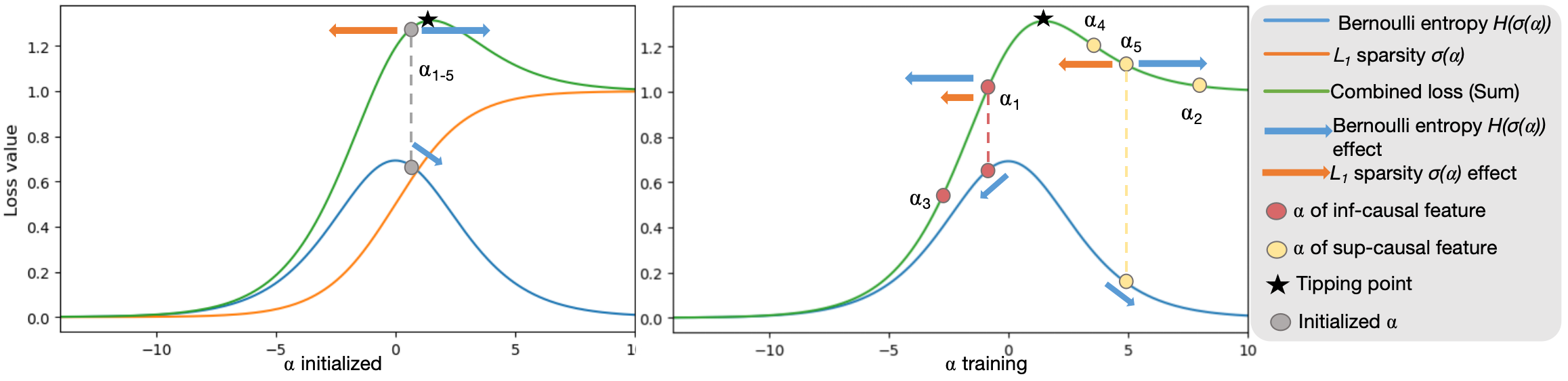}
  \caption{Conceptual diagram of the self-supervised Bernoulli regularization process. The figure illustrates the effect of the regularization on the evolution of $\alpha$ values for each feature immediately after initialization (left), contrasted with the effect on $\alpha$ values of features that have been partitioned after training (right). inf-causal: inferior causal, sup-causal: superior causal.}
  \label{fig:bernoulli_regularization}
\end{figure*}

We introduce a global, feature-wise gate that assigns each of the input features a latent Bernoulli activation variable $S_i$, where the probability $f_i=\sigma(\alpha_i)$ is a learnable parameter. This gate is designed to be sample-agnostic, providing consistent feature importance scores across domains. The gate's distribution is shaped by two complementary regularizers: i) an \textit{$\ell_1$ sparsity} term for encouraging overall sparsity, and ii) a \textit{Bernoulli entropy} term that polarizes importance scores, effectively performing a soft, differentiable feature selection. Coupled with the prediction task loss, this framework automatically partitions features into lower-importance (``inferior causal'') and higher-importance (``superior causal'') groups in a purely data-driven manner, as illustrated in Figure~\ref{fig:bernoulli_regularization}.

The global feature gate aims to classify the raw features into \textit{inferior causal features} (inf-causal) and \textit{superior causal features} (sup-causal), with both groups contributing to the representation but the latter exerting stronger influence.

To achieve this, we introduce a global gate parameter vector $\alpha\in\mathbb{R}^p$, with $p$ indicating the input dimension. 
A single, sample-agnostic weight vector $g$ is derived via a softmax activation:
\begin{equation}
\label{eq:gate}
    g \;=\; \mathrm{softmax}(\alpha)\in(0,1)^p,
    \quad \sum_{i=1}^p g_i=1
\end{equation}
Crucially, this gate vector $g$ is global, i.e., it is independent of the input sample $X$ and remains fixed across all instances and domains. This ensures that the feature importance is consistent, reflecting the intrinsic property of the feature itself rather than its specific value in a given context.
To prevent the gate from being influenced by the arbitrary scale of different features, we standardize each feature to a standard normal distribution. This ensures that the learned weights $g$ reflect genuine predictive importance rather than being correlated with feature magnitudes.

The following sections detail the design of the two complementary regularizers: the $\ell_1$ sparsity (for overall sparsity) and the Bernoulli entropy (for feature polarization).

\subsection{$\ell_1$ Sparsity}

To encourage a compact representation and suppress non-causal features, we apply an $\ell_1$ penalty on the feature importance scores. Specifically, we map the gate parameter $\alpha$ through a sigmoid function to obtain a vector of importance $f=\sigma(\alpha)\in(0,1)^p$. The $\ell_1$ sparsity term is defined as the sum:
\begin{equation}
\mathcal{L}_{\ell_1}
=\|f\|_1
=\sum_{i=1}^p \sigma(\alpha_i).
\label{eq:l1_sparsity}
\end{equation}

We adopt the sigmoid function here because the softmax output $g$ (from Eq.~\ref{eq:gate}) has a constant $\ell_1$ norm of 1 by construction, which would invalidate an $\ell_1$ penalty on $g$. The $\mathcal{L}_{\ell_1}$ penalty suppresses the weights of uninformative features, driving them toward zero. This process acts as an adaptive ``trial'' mechanism, which automatically identifies and down-weights features that are likely to be inf-causal.

\subsection{Bernoulli Entropy}

The $\ell_1$ penalty encourages sparsity but does not necessarily force features to be fully ON or OFF, leaving feature importance probabilities $f_i$ insufficiently distinct. To resolve this ambiguity and push the model toward a clear feature selection, we introduce Bernoulli entropy as the second regularizer of the importance distribution.

We define the Bernoulli probability for each feature as $f = \sigma(\alpha) \in (0,1)^p$, where $f_i = \Pr(S_i=1)$ for a hypothetical Bernoulli variable $S_i$. Note that this is a conceptual device used only to define the entropy, and no actual sampling is performed during training, ensuring the process remains differentiable.

Since there are no ground-truth labels indicating whether a feature is inf-causal or sup-causal, the partitioning (classification) must be learned in a \emph{self-supervised} manner. To this end, we minimize the sum of Bernoulli entropies (BE):
\begin{equation}
\mathcal{L}_{BE}
\;=\; \sum_{i=1}^p H\!\left(\mathrm{Bern}(f_i)\right)
\;=\; -\sum_{i=1}^p \Big[f_i\log f_i + (1-f_i)\log(1-f_i)\Big],
\label{eq:be}
\end{equation}
in which the loss reaches its maximum when $f_i=0.5$ and attains its minimum as $f_i$ approaches $0$ or $1$.

\paragraph{Theoretical justification.}
The polarizing effect of $\mathcal{L}_{BE}$ can be understood by analyzing its properties. For a single dimension, define the scalar entropy function $r(f) = H(\mathrm{Bern}(f))$ for $f \in (0,1)$. Its first and second derivatives are:
\[
r'(f)=\log\frac{1-f}{f}, 
\qquad 
r''(f)=-\frac{1}{f(1-f)}<0,
\]
The negative second derivative proves $r(f)$ is strictly concave, with a unique maximum at $f = \tfrac{1}{2}$ and minima at the boundaries $f \to 0$ and $f \to 1$.

In the multivariate case, the total loss $\mathcal{L}_{BE}(f) = \sum{i=1}^p r(f_i)$ is separable. Its Hessian is a diagonal matrix with entries $r''(f_i) < 0$, confirming that $\mathcal{L}_{BE}$ is strictly concave on $(0,1)^p$. The loss is uniquely maximized when all $f_i = 0.5$ and minimized only at the vertices of the hypercube, where $f \in {0,1}^p$.

Therefore, minimizing $\mathcal{L}_{BE}$ drives each $f_i$ away from ambiguity and toward a deterministic 0 or 1, effectively performing an \emph{unsupervised binary partition} of the features into \{inf-causal, sup-causal\}.

\subsection{Combined Regularization Dynamics}

The full regularization effect emerges from the combination of the $\ell_1$ sparsity and Bernoulli entropy terms. Together, they create a biased, bi-stable dynamic that pushes features toward either 0 or 1, with a tipping point determined by the ratio of the regularization strengths. The task loss then determines which features cross this threshold, resulting in a self-supervised partition into inferior and superior causal groups.

The combined regularizer is defined as:
\begin{equation}
\mathcal{L}_{\ell_1BE}(f)
\;=\;
\lambda_{BE} \underbrace{\sum_{i=1}^p H\!\left(\mathrm{Bern}(f_i)\right)}_{\mathcal{L}_{BE}}
\;+\;
\lambda_{\ell_1} \underbrace{\sum_{i=1}^p f_i}_{\mathcal{L}_{\ell_1}},
\label{eq:comb_reg}
\end{equation}
where $\mathcal{L}_{BE}$ encourages polarization ($f_i \to \{0,1\}$) and $\mathcal{L}_{\ell_1}$ penalizes overall activation mass.

\paragraph{Initialization and symmetry breaking.}
To break the symmetry inherent in random initialization (where $f_i \approx \tfrac12$), we initialize all gate parameters $\alpha$ with a small positive bias (e.g., $\alpha=0.01$). This results in initial probabilities $f_i=\sigma(0.01)\approx 0.5025$, providing a uniform but deterministic starting point slightly above $0.5$, which avoids random initial flips and ensures stable optimization.

\paragraph{Gradient dynamics and the tipping point.}
By the chain rule, the gradient of the combined regularizer with respect to the gate parameters is:
\[
\frac{\partial \mathcal{L}_{\ell_1BE}}{\partial \alpha_i}
=
\frac{\partial \mathcal{L}_{\ell_1BE}}{\partial f_i}
\cdot
\frac{\partial f_i}{\partial \alpha_i}
\;=\;
\underbrace{\Big(\lambda_{BE} \log\frac{1-f_i}{f_i} + \lambda_{\ell_1}\Big)}_{B(f_i)}\,\underbrace{f_i(1-f_i)}_{>0}.
\]
Hence, the sign of the gradient is governed entirely by the bracket $B(f_i)$. Setting $B(f_i)=0$ yields a \emph{tipping point}:
\begin{equation}
f_i^\star
\;=\;
\frac{1}{1+\exp(-\lambda_{\ell_1}/\lambda_{BE})}
\;=\;
\sigma\!\left(\frac{\lambda_{\ell_1}}{\lambda_{BE}}\right)
\;>\;\tfrac12.
\label{eq:tipping}
\end{equation}
Because $f_i(1-f_i)\!>\!0$, this tipping point defines the gradient-descent dynamics:
\[
\begin{cases}
f_i < f_i^\star \quad \Rightarrow \quad \frac{\partial \mathcal{L}_{\ell_1BE}}{\partial \alpha_i} > 0 \;\Rightarrow\; \alpha_i \downarrow \;\Rightarrow\; f_i \downarrow \;\to 0,\\[2mm]
f_i > f_i^\star \quad \Rightarrow \quad \frac{\partial \mathcal{L}_{\ell_1BE}}{\partial \alpha_i} < 0 \;\Rightarrow\; \alpha_i \uparrow \;\Rightarrow\; f_i \uparrow \;\to 1.
\end{cases}
\]
Thus, the $\mathcal{L}_{\ell_1}$ term shifts the unstable equilibrium from $0.5$ (pure Bernoulli entropy) to $f_i^\star>0.5$, creating a bias where features must ``prove their worth'' to be activated.

\paragraph{Interaction of the regularizers.}
The two regularizers interact synergistically: i) The Bernoulli entropy term ($\mathcal{L}_{BE}$) acts as a polarizing force, pushing $f_i$ away from $0.5$ toward the extremes; ii) the $\ell_1$ sparsity term ($\mathcal{L}_{\ell_1}$) acts as a global suppressing force, providing a constant positive bias in the gradient that raises the activation threshold to $f_i^\star$. This makes it harder for a feature to remain active unless it is strongly beneficial for the task.

\paragraph{Self-supervised selection.}

The combined effect, in concert with the task loss, facilitates a self-supervised selection process:
\begin{itemize}[leftmargin=*]
\item Features that provide strong predictive signals receive gradients from the task loss that help them overcome the raised threshold $f_i^\star$. Once above, the entropy term pushes them further toward 1, cementing them as sup-causal features.
\item Features that provide weak signals are unable to overcome the $\ell_1$ suppression. They fall below $f_i^\star$ and are pushed by both regularizers toward 0, becoming inf-causal features.
\end{itemize}
In this way, the model performs an unsupervised binary feature partition guided by the target prediction objective.

\section{Domain-Wise Training}

To learn representations that generalize across domains, we train across multiple domains by aggregating per-domain losses with difficulty-aware weights and by adding regularizers that align performance and disentangle causal representation. This ensures the model learns succinct invariant mechanisms rather than domain-specific shortcuts. Its contribution is further substantiated through the ablation study comparing LSTM and Cabernet-SIRM in Section~\ref{sec:ablation_reg}.

\subsection{Difficulty-aware Aggregation}

A simple average of per-domain losses can be biased towards larger datasets or domains with inherently higher-variance signals, and thus cause the model to overfit to these dominant distributions. To overcome this, we minimize a difficulty-weighted average of per-domain losses, which aims to balance the contribution of each domain regardless of its size or inherent complexity.

Specifically, for each domain $e$ in a mini-batch (with samples from multiple domains), we compute its domain-specific NMSE:
\[
L_e \;=\; \operatorname{NMSE}\!\big(m_\theta(h_\phi(X^e)),\, Y^e\big).
\]
We then aggregate these losses into a global target using difficulty-based weights:

\begin{equation}
\mathcal{L}_{target} \;=\; \frac{1}{|\mathcal{E}|}\sum_{e\in\mathcal{E}} w_e\,L_e,
\end{equation}
where the weights $w_e$ are designed to counterbalance domain difficulty. We estimate this difficulty score $d_e$ by combining the coefficient of variation (CV) of the target and the mean CV of the inputs:
\[
\text{CV}_Y^{(e)} = \frac{\operatorname{Std}(Y^e)}{\mathbb{E}[\lvert Y^e\rvert]},
\text{CV}_X^{(e)} = \frac{1}{p}\sum_{j=1}^p 
\frac{\operatorname{Std}(X^e_{\cdot j})}{\mathbb{E}[\lvert X^e_{\cdot j}\rvert]},
d_e = \tfrac{1}{2}\big(\text{CV}_Y^{(e)}+\text{CV}_X^{(e)}\big).
\]
The weight for each domain is then set as the inverse of its difficulty score: $w_e = 1/d_e$. This scheme prevents any single complex or large domain from dominating the objective.

\subsection{Cross-domain Variance Regularization}
Motivated by the Causal Invariance Assumption (Guideline 1, Section~\ref{para:cia}), we penalize the dispersion of difficulty-weighted domain losses within a batch, nudging the representation toward causal invariance rather than domain-specific fits.

We encourage $h_\phi$ to capture relationships that are consistently predictive across all training domains, rather than specialized, domain-specific shortcuts. Concretely, we penalize the dispersion of the difficulty-weighted per-domain losses within each batch:
\begin{equation}
\mathcal{L}_{var}
\;=\;
\frac{1}{|\mathcal{E}|}\sum_{e\in\mathcal{E}} \big(w_e L_e - \bar{L}\big)^2 .
\end{equation}
Driving $\mathcal{L}_{var}$ down nudges the representation toward a mechanism that performs consistently across domains, aligning with the \emph{causal invariance} assumption that the conditional $P(Y \mid X_{S^*})$ should be stable over environments. This encourages the model to capture causal representations rather than domain-specific correlations.

\subsection{Independence of Causal Representation}
In accordance with the Independence of Causal Representation Assumption (Guideline 3, Section~\ref{par:indy}), we encourage the learned latent representation $Z$ to be composed of statistically independent factors. This promotes disentanglement, improves interpretability, and prevents the model from learning redundant, entangled representations.

Let $Z = h_\phi(X) \in \mathbb{R}^d$ be the latent representation for a batch of data. We compute the sample correlation matrix $C$ of the latent dimensions across the batch:
\[
C_{ij} \;=\; \frac{\langle Z_{i}, \, Z_{j} \rangle}
{\|Z_{i}\|\;\|Z_{j}\|},
\quad \text{for} \,\,  i,j = 1,2,\ldots,d,
\]
where $Z_{i}$ denotes the $i$-th dimension of $Z$.

We then promote independence by shrinking the off-diagonals of $C$, penalizing the deviation from the identity:
\begin{equation}
\label{eq:L_indy}
\mathcal{L}_{indy}
\;=\;
\frac{1}{d(d-1)}\|C - I_d\|_1
\;=\;
\frac{1}{d(d-1)} \sum_{i\neq j} \lvert C_{ij}\rvert,
\end{equation}
where $d(d-1)$ serves as a normalization factor. 

This operationalizes the Independence of Causal Representation assumption and discourages redundant entanglement among latent coordinates.

\section{Experiments}

\begin{figure}[t]
  \centering
  \includegraphics[width=0.99\linewidth]{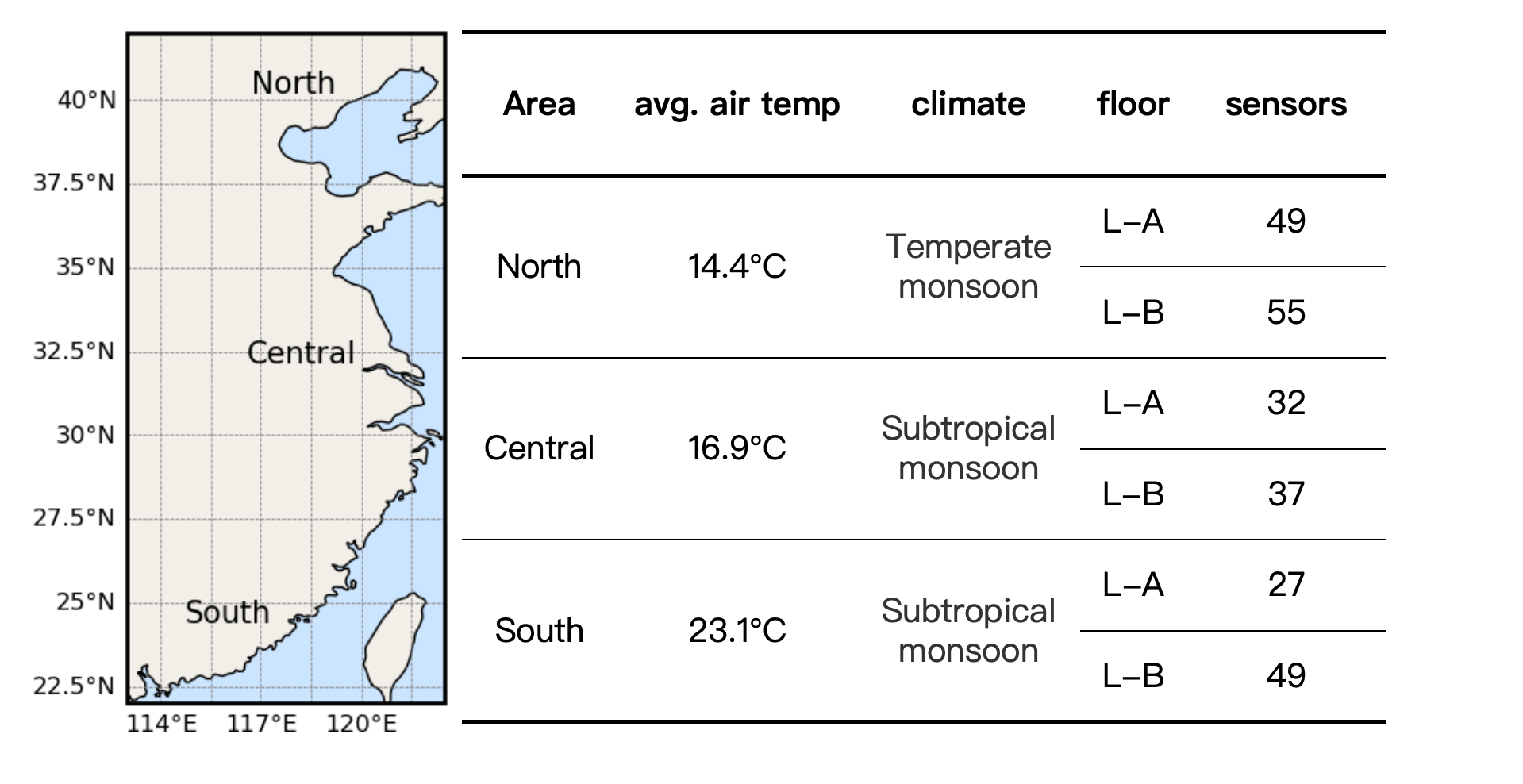}
  \caption{Dataset summary and collection sites.}
  \label{fig:ds}
  \vspace{-10pt}
\end{figure}

\textbf{Datasets.} Publicly available datasets that combine indoor sensing with HVAC energy consumption across multiple climate zones remain scarce. Therefore, our experiments are conducted on data we collected via multi-site sensor deployments. The data comprises six office-floor datasets from three commercial buildings located in three Chinese cities, spanning a north–Central–south climate gradient (Figure~\ref{fig:ds}). 
Indoor sensing is deployed uniformly across the occupied areas. 
Per timestamp, the feature set includes: indoor temperature, indoor humidity, light intensity, \(\mathrm{CO_2}\) concentration, indoor air pressure, and total volatile organic compounds (TVOC). 
Indoor sensors log at a 5-minute resolution. 
We also ingest outdoor temperature collected from NASA POWER~\cite{nasa_power} and resample it to 5-minute resolution via linear interpolation to match the indoor data cadence. The data used in the experiments cover the period from November 2024 to July 2025.
\\
\textbf{Data processing and feature engineering.}
A complete pipeline is provided in our code.  
Because different floors may host varying numbers of devices (Figure~\ref{fig:ds}), we aggregate features using fixed summary statistics, ensuring consistent input dimensionality per dataset, regardless of the device count. 
To handle missing data, we apply a standardized pipeline (details in our code) for alignment and short-gap filling, while discarding segments that fail integrity checks.
We add a binary feature \(\texttt{is\_work}\) that marks work hours/days (including make-up days) versus off-hours/holidays.
Besides, for each domain, we compute a summary of the energy scale (see Section~\ref{sec:energy_predictor}). 
The z-score normalization is applied to features.
Finally, we window the multivariate time series into length-12 input sequences (the most recent 60 minutes) and predict the next 5-minute AC energy consumption.
\\
\textbf{Evaluation protocol.}\label{sec:protocol}
To remove the effect of scale differences across domains, we report the normalized mean squared error (NMSE). 
For a test domain $e$ with ground-truth $\{y_t^e\}_{t=1}^{n_e}$ and predictions $\{\hat{y}_t^e\}_{t=1}^{n_e}$; we define
\[
\mathrm{NMSE}^e 
\;=\;
\frac{\sum_{t=1}^{n_e}\big(y_t^e-\hat{y}_t^e\big)^2}
     {\sum_{t=1}^{n_e}\big(y_t^e)^2}.
\]
We adopt a leave-one-domain-out protocol, which is standard in domain generalization~\cite{li2017deeper}: 
Each time, one domain is held out as a test, and the remaining domains form the training pool. 
Within every training domain, we split $20\%$ of data as validation and use the remaining $80\%$ for fitting; the model checkpoint achieving the best validation NMSE is then evaluated on the held-out test domain. 
\\
\textbf{Implementation details.}
We train all models with Adam optimizer (learning rate $2\times10^{-4}$) for 500 epochs. 
Unlike many of the domain generalization literatures~\cite{lv2022causality,liu2024causality}, we avoid any data augmentation or causal intervention procedures, as they may inject human-induced biases. Our approach aims to remain as data-driven as possible, relying on the training distributions rather than external knowledge.
The hidden size is $d=64$ and the batch size is $512$. 
All experiments are conducted on a NVIDIA RTX A6000 GPU.

\subsection{Performance Comparison}
\label{sec:comparison}
We compare against four representative methods. LSTM~\cite{hochreiter1997long} serves as a classical deep sequence model and backbone of the CaberNet. LSTM–LiNGAM combines causal discovery with sequence prediction: we first estimate a causal graph from training data using direct-LiNGAM~\cite{shimizu2011directlingam}, extract the Markov blanket of $Y$ from the discovered graph, and then train an LSTM using only the blanket variables for prediction. 
We also include two recent benchmarks tailored to HVAC energy prediction: 
Shift-GRU~\cite{liu2023energy}, a GRU regressor trained on inputs pre-aligned by per-feature time lags estimated via Pearson correlation; and 
STL~\cite{xing2024transfer}, a transfer-learning LSTM that trains on samples most similar to the target domain.
\sysname{} (Ours) integrates causal representation learning with sequence modeling. All models are trained using the same protocol (see Section~\ref{sec:protocol}). The result is shown in Table~\ref{tab:nmse_results}.
We mainly compare with the LSTM-based approach since it also serves as the backbone of \sysname{}, ensuring a fair and controlled evaluation.

\begin{table}[h]
  \caption{Cross Domain Evaluation (\textbf{bold}: best, \underline{underline}: second, \textit{italic}: equal)}
  \centering
  \footnotesize
  \begin{tabular}{lccccccc}
    \toprule
     & \multicolumn{7}{c}{NMSE} \\
    \cmidrule(lr){2-8} 
    Method & \multicolumn{2}{c}{North} & \multicolumn{2}{c}{Central} & \multicolumn{2}{c}{South} & \multirow{2}{*}{Average}\\
    \cmidrule(lr){2-3} \cmidrule(lr){4-5} \cmidrule(lr){6-7} 
    & L-A & L-B & L-A & L-B & L-A & L-B \\
    \midrule
    LSTM & 0.114 & \underline{0.130} & \underline{0.165} & \textit{0.286} & 0.368 & 0.354 & 0.236\\
    LSTM-LiNGAM & \underline{0.064} & 0.198 & 0.206 & \textit{0.286} & 0.381 & 0.318 & 0.242\\
    Shift-GRU & 0.139 & 0.190 & 0.189 & 0.291 & \underline{0.289} & \underline{0.287} & \underline{0.231}\\
    STL & 0.114 & 0.161 & 0.213 & 0.363 & 0.356 & 0.334 & 0.257\\
    \textbf{\sysname{}}  & \textbf{0.063} & \textbf{0.119} & \textbf{0.131} & \textbf{0.254} & \textbf{0.261} & \textbf{0.242} & \textbf{0.178}\\
    \bottomrule
  \end{tabular}
  
  \label{tab:nmse_results}
\end{table}
\sysname{} achieves the best NMSE on all six domains (Table~\ref{tab:nmse_results}); averaged over domains it improves by \textbf{22.9\%} relative to the best benchmark. 
Per-domain gains over other benchmarks are consistent and substantial, indicating that causal-invariance regularization and soft feature selection improve generalization across buildings and climates. Moreover, \sysname{} is more interpretable (see Section~\ref{sec:explainability}), as its causal representation is both transparent in composition and grounded in causal reasoning. This transparency supports industrial deployment and operator trust, which is especially important in human health–related HVAC applications.

Shift-GRU achieves the second average NMSE and is especially strong in the South, where it beats plain LSTM, likely because per-feature time-lag alignment mitigates phase shifts from thermal inertia and lag offsets in warm-humid climates. However, Pearson correlation is linear and pairwise, so its shifts can be unreliable under nonlinear or multi-lag dynamics. This limits robustness relative to \sysname{}. By contrast, STL performs worst overall: its similarity-based filtering shrinks data diversity and sample size, which is ill-suited when data are scarce and highly variable across domains.

LSTM–LiNGAM is overall comparable to LSTM, and in some cases markedly better (North L-A and South L-B), probably because the discovered blanket retained strong drivers (e.g., \textit{outdoor\_temperature}, \textit{is\_work}) while discarding weak ones, as shown in Appendix~\ref{app:lingam}. 
However, its performance varies: hard masking can remove useful proxy variables; LiNGAM also assumes a linear, acyclic data-generating process with non-Gaussian errors and no latent confounding~\cite{shimizu2011directlingam}, which may be too restrictive for the complex and nonlinear nature of HVAC dynamics. 
Moreover, LiNGAM’s binary include/exclude decision is brittle: features that are not strictly causal but still carry a latent causal signal (e.g., proxies, mediators, regime indicators) are dropped entirely. By contrast, \sysname{} uses soft gating to retain such partial contributions while appropriately down-weighting them. In this way, the model can still exploit these signals in forming the final causal representation, thereby improving robustness under domain shift.
These misspecifications can yield suboptimal blankets and underperformance on other floors. 
Notably, in Central L-B LiNGAM selected essentially the full feature set as the blanket, making its input identical to LSTM and resulting in identical NMSE. The detailed SCMs constructed by LiNGAM are shown in Appendix~\ref{app:lingam}.
\subsection{Ablation Study}

\begin{figure*}[h]
  \centering
  \includegraphics[width=0.85\linewidth]{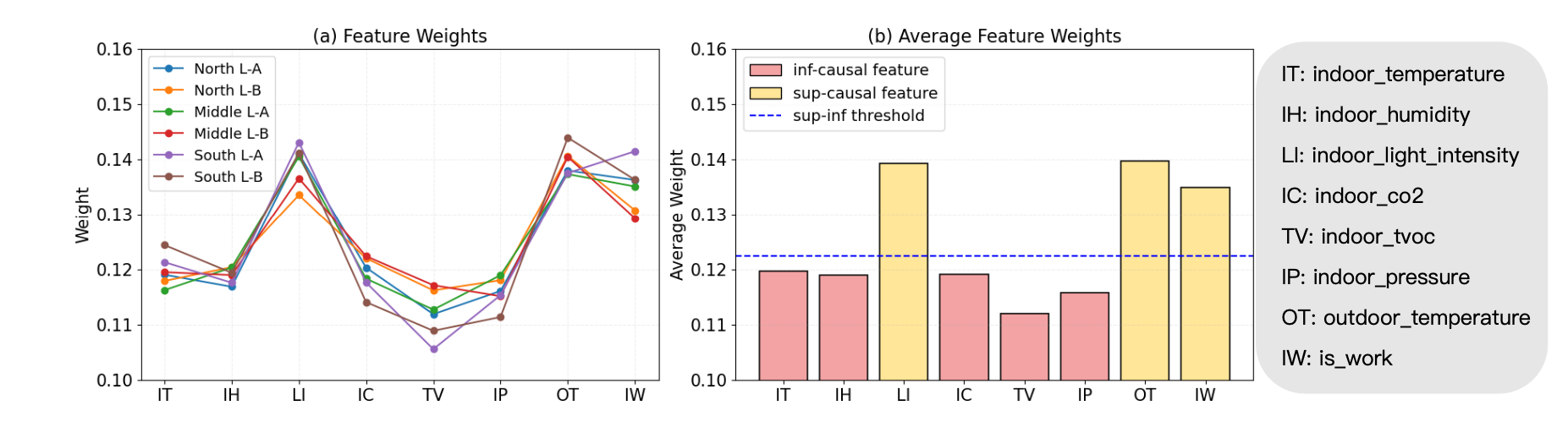}
  \caption{Feature weights (left: test on labeled dataset, right: average over all datasets).}
  \label{fig:fw_both}
  \vspace{-5pt}
\end{figure*}

\subsubsection{Regularization}
\label{sec:ablation_reg}
We ablate the Bernoulli regularization $\mathcal{L}_{\ell_1BE}$ by removing the global feature gate and also dropping the cross-domain variance and independence penalties. 
The remaining objective aggregates losses \emph{per domain} (equal domain contribution) with the difficulty coefficient on the target loss only. 
This ablation is called \sysname{}-SIRM (simplified invariant risk minimization), which balances risk across environments, but without an explicit invariance regularization; it stands between plain ERM (the LSTM baseline in Section~\ref{sec:comparison}) and our full \sysname{} model.

\begin{table}[h]
  \caption{Ablation Results on Regularization
  }
  \centering
  \footnotesize
  \begin{tabular}{lccccccc}
    \toprule
     & \multicolumn{7}{c}{NMSE} \\
    \cmidrule(lr){2-8} 
    Method & \multicolumn{2}{c}{North} & \multicolumn{2}{c}{Central} & \multicolumn{2}{c}{South} & \multirow{2}{*}{Average}\\
    \cmidrule(lr){2-3} \cmidrule(lr){4-5} \cmidrule(lr){6-7} 
    & L-A & L-B & L-A & L-B & L-A & L-B \\
    \midrule
    \sysname{}-SIRM & 0.075 & 0.162 & 0.134 & 0.294 & 0.301 & 0.268 & 0.206\\
    \sysname{}  & \textbf{0.063} & \textbf{0.119} & \textbf{0.131} & \textbf{0.254} & \textbf{0.261} & \textbf{0.242} & \textbf{0.178}\\
    \bottomrule
  \end{tabular}
  \label{tab:nmse_ablation}
\end{table}

\textbf{LSTM (ERM) vs. \sysname{}-SIRM (IRM)}. 
Compared to the LSTM baseline (Table~\ref{tab:nmse_results}), ERM can occasionally win on a specific test domain (e.g., \emph{North L-B}). 
This happens because ERM’s sample-proportional training implicitly upweights large/source-like domains; when such a domain is “lucky” and happens to resemble the held-out target distribution, ERM benefits from this alignment. 
By contrast, \sysname{}-SIRM gives each training domain equal influence, foregoing that accidental advantage. 
This ``ERM$>$IRM'' phenomenon is common, especially when ERM is well tuned~\cite{rosenfeld2020risks,gulrajani2020search}; nonetheless, \emph{on average} \sysname{}-SIRM still outperforms ERM by \textbf{12.7\%}, indicating that equalizing domain contributions reduces overfitting to domain-specific variations of large domains and improves robustness across diverse targets.

\textbf{\sysname{} vs. \sysname{}-SIRM}.
We see \sysname{} has consistent additional gains, yielding a \textbf{13.6\%} relative improvement.
The full model’s advantages come from 
(i) \emph{Bernoulli regularization}: the combined $\mathcal{L}_{\ell_1BE}$ objective polarizes gate probabilities toward $\{0,1\}$ and raises the neutrality threshold $f^\star$ (Eq.~\ref{eq:tipping}), suppressing weak spurious cues while preserving genuinely predictive signals, thereby yielding stable, parsimonious inputs to $m_\theta$;
(ii) the cross-domain variance penalty $\mathcal{L}_{var}$, which steers optimization toward invariant mechanisms rather than domain-specific correlations.
Together, these elements deliver the best cross-domain generalization.
\subsubsection{Independence of causal representation}
\label{sec:ablation_indy}
We ablate the independence loss $\mathcal{L}_{indy}$ (Eq.\ref{eq:L_indy}), thereby not enforcing the \emph{Independence of Causal Representation} hypothesis (Guideline 3, Section\ref{par:indy}), while keeping all other components identical.
We refer to this variant as \sysname{}–NoIndy. 
To assess compactness, we vary the latent size $d\!\in\!\{32,16,8\}$ (see Tables~\ref{tab:avg_nmse_hiddensize} and details in Appendix \ref{app:indy}).

\begin{table}[h]
  \caption{Ablation Results on Independence of $Z$}
  \centering
  \footnotesize
  \begin{tabular}{lccc}
    \toprule
    \multirow{2}{*}{Method} & \multicolumn{3}{c}{Hidden Size} \\
    \cmidrule(lr){2-4} 
     & 8 & 16 & 32 \\
    \midrule
    \sysname{}-NoIndy  & 0.237 & 0.232 & 0.207\\
    \sysname{} & \textbf{0.214} & \textbf{0.210} & \textbf{0.192}\\
    \bottomrule
  \end{tabular}
  \label{tab:avg_nmse_hiddensize}
\end{table}

Overall, \sysname{} outperforms \sysname{}-NoIndy. 
The independence loss compresses redundancy and allocates capacity to informative factors, so the latent dimensions carries useful and non-overlapping signal. 
Consequently, when reducing model width, \sysname{} degrades more slowly: from hidden size $32\!\rightarrow\!16$ the relative NMSE increase is about \textbf{9.4\%} for \sysname{} versus 12.1\% for \sysname{}-noIndy; from $32\!\rightarrow\!8$ it is about \textbf{11.5\%} versus 14.5\%, respectively. 
This stability suggests that encouraging factor independence reduces the redundancy in the representation, making the model less sensitive to width constraints. This is consistent with prior research on independence-promoting bottlenecks~\cite{lv2022causality}.

\subsection{Explainability}
\label{sec:explainability}

\subsubsection{Feature weighting}
\label{sec:fw}

We extract global gate weights $g$ (Eq.~\ref{eq:gate}) that reweight standardized raw features before they enter the causal representation extractor. 
Rather than hard-masking features as in LiNGAM-style selection, the gate performs \emph{soft} selection: it assigns continuous importances and yields a data-driven split into \emph{sup-causal} (higher weight) and \emph{inf-causal} (lower weight) variables. 
Softness is desirable because even features with weaker direct effects can carry a latent causal signal or act as proxies that help form a better causal representation $Z$; they are therefore down-weighted rather than discarded.


\textbf{Per-domain patterns (Figure~\ref{fig:fw_both} left).}  
Across all six domains, the curves exhibit a highly similar ranking of feature importance with only mild dispersion. 
This cross-domain alignment suggests that the gate captures relationships that are stable rather than domain-specific, consistent with the \emph{causal invariance} assumption (Section~\ref{para:cia}). 
In other words, the learned importance profile transfers across buildings/floors/cities, which supports our goal of domain-robust prediction.


\textbf{Averaged importance (Figure~\ref{fig:fw_both} right).}  
Two variables, namely \textit{TVOC} and \textit{indoor air pressure}, consistently fall below the sup/inf threshold and are classified as \emph{inf-causal}; this is plausible since they do not directly drive AC operation in these offices.
\textit{Outdoor temperature} receives the largest weight, which agrees with building physics and AC control logic. 
The calendar proxy \textit{is\_work} is also strongly weighted, reflecting occupancy-driven operation typical of office buildings. 
\textit{Light intensity} appears higher than one might expect if viewed only as a heat-gain channel; in practice it functions as a robust proxy for \emph{time-of-day/usage regime} (natural light cycles and artificial lighting schedules), which is highly informative for AC operation in the absence of an explicit time feature, and thus effectively following the pathway: 
$
\text{light} \to \text{time} \to \text{AC operation}
$.

By contrast, \textit{indoor temperature} and \textit{indoor humidity} have relatively modest weights: when AC is on, these variables are actively regulated near setpoints that meet human comfort (low variability, limited incremental predictive signal), and when AC is off (e.g., nights), their drift toward outdoor conditions does not immediately translate into consumption since the AC is not operating. Thus, they provide limited incremental predictive value.


\subsubsection{Jacobian-Based Feature Contribution}
\label{sec:jacobian}
Motivated by previous work~\cite{zhang2024opti,selvaraju2016grad}, we quantify how each raw feature contributes to each dimension of the $d$-dimensional $Z$ using Jacobians and then \emph{de-bias} magnitudes by removing the scaling effect of the global gate's weights. For clarity, we fix the hidden size to $d=5$.

\textbf{Jacobian-based contribution.}
Let the input window be $X_{t-w:t-1}\in\mathbb{R}^{w\times p}$ with $p$ raw features per time step and $w$ steps, and let the learned representation be $Z\in\mathbb{R}^d$. We define the time-averaged Jacobian (magnitude) with respect to the \emph{original} inputs as
\[
J
\;=\;
\frac{1}{w}\sum_{t=1}^{w} \Bigl|\tfrac{\partial Z}{\partial X_t}\Bigr|
\;\in\mathbb{R}^{d\times p},
\]
whose $(i,j)$ entry measures the absolute influence of the $j$-th raw feature on $Z_i$ across the time window.

To eliminate the confounding influence of the gate weights, we debias the contribution matrix by right-multiplication with $\mathrm{diag}(g)^{-1}$; using the chain rule (see Appendix~\ref{app:debias-deriv}),
\begin{equation}
\label{eq:J_deb}
J_{\text{debiased}}
\;=\;
J\,\cdot\mathrm{diag}(g)^{-1}
\
\end{equation}

\textbf{Why debias?}
$J$ reflects the dependency actually used by the trained model (post-gating), while $J_{\text{debiased}}$ reveals the intrinsic sensitivity of the encoder absent the structural rescaling by $g$. 
A large $J_{\text{debiased}}$ but small $J$ indicates a feature that is intrinsically informative yet suppressed by the gate, whereas large values in both $J_{\text{debiased}}$ and $J$ indicate a feature that is both informative and actively used by the model. This bias is discussed deeply in work~\cite{wang2024gradient}.

\begin{table}[t]
\centering
\caption{Top-3 Contributing Features per Representation Dimension (Original and Debiased Jacobian)}
\label{tab:top3}
\footnotesize
\begin{tabular}{lccc}
\toprule
\multicolumn{4}{c}{\textbf{Original Jacobian}} \\
\midrule
$Z$ & 1st Feature & 2nd Feature & 3rd Feature \\
\midrule
$Z_0$ & outdoor\_temperature & indoor\_light & indoor\_pressure \\
$Z_1$ & outdoor\_temperature & indoor\_pressure & is\_work \\
$Z_2$ & indoor\_pressure & outdoor\_temperature & indoor\_temperature \\
$Z_3$ & outdoor\_temperature & indoor\_light & is\_work \\
$Z_4$ & indoor\_humidity & outdoor\_temperature & indoor\_pressure \\
\midrule
\multicolumn{4}{c}{\textbf{Debiased Jacobian}} \\
\midrule
$Z_0$ & outdoor\_temperature & indoor\_light & indoor\_pressure \\
$Z_1$ & indoor\_pressure & outdoor\_temperature & indoor\_tvoc \\
$Z_2$ & indoor\_pressure & outdoor\_temperature & indoor\_temperature \\
$Z_3$ & outdoor\_temperature & indoor\_light & is\_work \\
$Z_4$ & indoor\_humidity & outdoor\_temperature & indoor\_pressure \\
\bottomrule
\end{tabular}
\end{table}

The top-3 contributing features per $Z_i$ are summarized in Table~\ref{tab:top3}, and the full numerical values are provided in Appendix~\ref{app:contrib}. We summarize the key observations below.

\textbf{(1) Sup-/inf-causal alignment.} According to Section~\ref{sec:fw}, we regard \textit{outdoor\_temperature}, \textit{indoor\_light}, and \textit{is\_work} as sup-causal. In Table~\ref{tab:top3}, these three sup-causal features account for \textbf{9/15} of the Top-3 slots under the original $J$ and \textbf{8/15} under the debiased $J_{\text{debiased}}$. By contrast, the smallest-weight inf-causal feature \textit{indoor\_tvoc} appears only \textbf{once} (and only in the debiased view). This pattern is consistent with the gate-derived sup/inf split: sup-causal features contribute more to the learned causal representation.

\textbf{(2) Dominance of outdoor temperature with complementary specialization.} Although its magnitude shrinks after debiasing, \textit{outdoor\_temperature} remains within the Top-2 for \emph{every} $Z_i$ in both $J$ and $J_{\text{debiased}}$, reflecting its primacy as a physical driver (also see its highest average weight in Figure~\ref{fig:fw_both}). The other Top-3 entries vary across $Z_i$, suggesting complementary specialization among latent coordinates: the model preserves the dominant outdoor signal while, under the independence regularizer $\mathcal{L}_{indy}$ (Eq.~\ref{eq:L_indy}), allocating the remaining capacity to less redundant features so as to balance prediction accuracy and $\mathcal{L}_{indy}$.

\textbf{(3) Role of \textit{indoor\_light} and \textit{is\_work}.} These two sup-causal proxies appear fewer times than \textit{outdoor\_temperature} in Top-3 lists, which is consistent with their indirect action on energy (time-of-day/usage regime indicators) discussed in Section~\ref{sec:fw}. They are likely distributed across multiple $Z_i$ and partially decomposed, rather than concentrated in a few coordinates. Importantly, Table~\ref{tab:contrib} in Appendix shows that they maintain high–mid ranks overall, even not often in the Top-3, surpassing most inf-causal features.

\balance

\subsubsection{Causal Discovery: SCM Reconstruction}
\label{sec:cd}

\begin{figure}[t]
  \centering
  \includegraphics[width=0.65\linewidth]{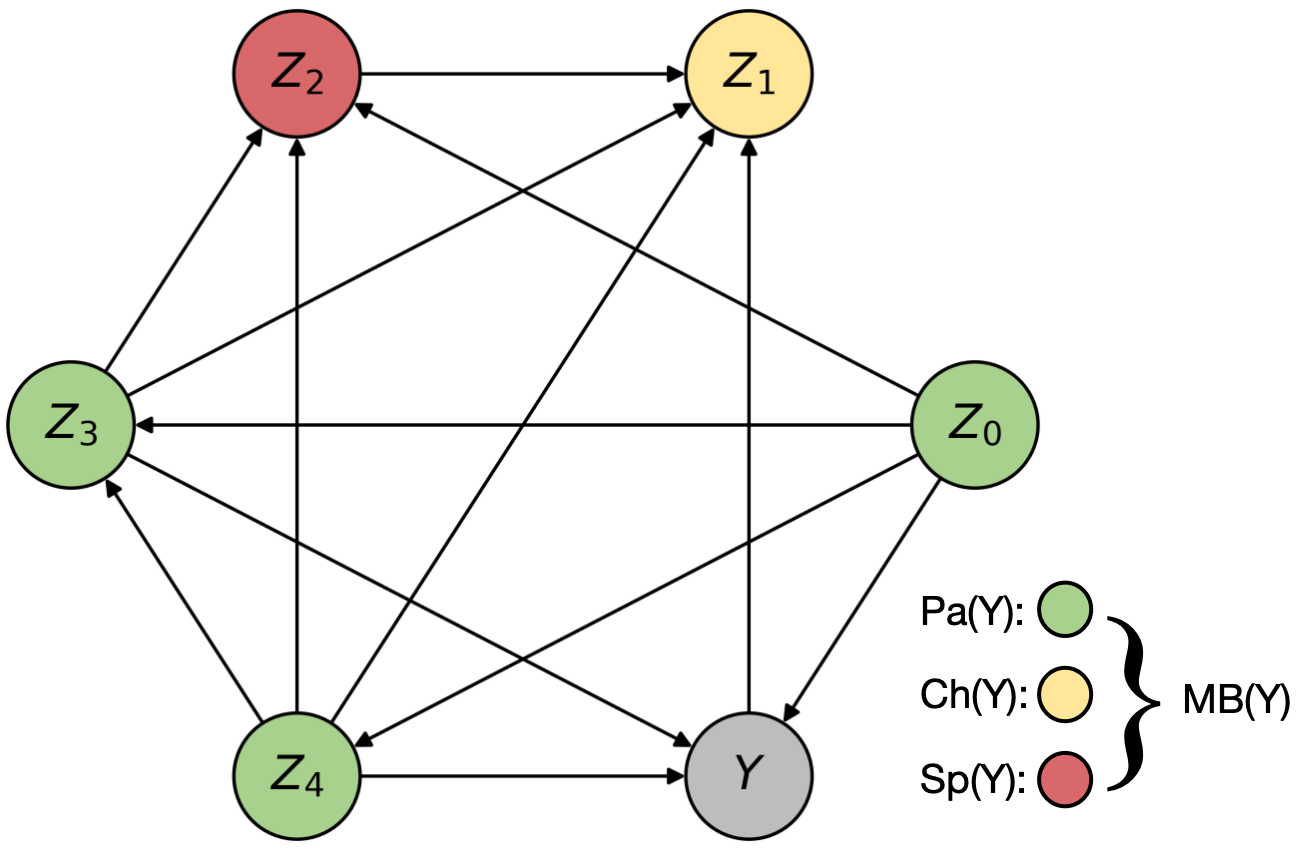}
  \caption{Reconstructed SCM (hidden size: 5).}
  \label{fig:mkb_z}
\end{figure}

Based on Section~\ref{sec:jacobian}, we reconstruct an SCM over the latent coordinates $Z$ and energy consumption $Y$. We use the same direct-LiNGAM~\cite{shimizu2011directlingam} algorithm as in Section~\ref{sec:comparison} to model causality. The reconstructed SCM is shown in Figure~\ref{fig:mkb_z}. We next interpret the reconstructed SCM using the contribution matrix above (Table~\ref{tab:top3}).

\textbf{(1) Markov blanket alignment.} $Z$ accurately represents the Markov blanket of $Y$ (Figure~\ref{fig:mkb_z}), aligning with our objective that the causal representation extractor $h_\phi$ learns Markov blanket representations (Section~\ref{sec:cre}) and the \emph{Guideline 2: Markov Blanket Prediction} (Section~\ref{para:g2mb}). 

\textbf{(2) High in-degree aggregators.} In Figure~\ref{fig:mkb_z}, $Z_1$ and $Z_2$ exhibit higher in-degree (multiple incoming arrows), indicating they act as aggregators of upstream signals. Consistently, their debiased Jacobian $J_{\text{debiased}}$ is topped by indoor variables (Table~\ref{tab:top3}; \textit{indoor\_pressure} for both), which aligns with the view that they are primarily \emph{affected} nodes rather than pure drivers.

\textbf{(3) Indoor thermodynamic response captured by $Z_2$.} Among all coordinates, \textit{indoor\_temperature} appears in the Top-3 \emph{only} for $Z_2$ (both $J$ and $J_{\text{debiased}}$), suggesting that $Z_2$ encodes more the building’s indoor thermal response. The reconstructed SCM shows arrows from other latents into $Z_2$; notably, $Z_0$, $Z_3$, and $Z_4$ are strongly associated with \textit{outdoor\_temperature} in Table~\ref{tab:top3}. This is consistent with the physical pathway
\[
\text{outdoor} \;\rightarrow\; \text{building dynamics} \;\rightarrow\; \text{indoor},
\]
i.e., outdoor-driven temperature latents feeding into the indoor-response temperature latent.

\textbf{(4) Parents of $Y$.} Edges into $Y$ originate from latents whose Top-3 include \textit{outdoor\_temperature} and \textit{indoor\_humidity}/\textit{pressure} (e.g., $Z_0$, $Z_3$, $Z_4$; see Table~\ref{tab:top3}), which is plausible: outdoor thermal load and indoor moisture/airmass proxies are primary physical drivers of AC energy. Latents associated with regime proxies (\textit{indoor\_light}, \textit{is\_work}; e.g., $Z_3$) tend to contribute either directly or indirectly, matching their proxy nature discussed in Section~\ref{sec:fw}.

\section{Conclusion}

We presented \sysname{}, a causal and interpretable representation learning framework for cross-domain energy prediction. The method couples a global feature gate with self-supervised Bernoulli regularization and domain-wise objectives for bias learning toward invariant, mechanism-level relations. Evaluated on real-world datasets collected from buildings in different climate zones, \sysname{} achieves state-of-the-art out-of-domain performance, improving average NMSE by 22.9\% over the best benchmark. Moreover, its interpretable latent representations align with domain knowledge, providing transparency critical for real-world energy optimization. Overall, \sysname{} overcomes a fundamental limitation in current approaches to cross-domain HVAC energy prediction and helps develop more adaptive, efficient, and automated HVAC control strategies across diverse built environments.

\newpage
\bibliographystyle{ACM-Reference-Format}
\bibliography{sample-base}

\appendix
\onecolumn

\section{Proof of Equation: Lower Conditional Variance with the Markov Blanket}
\label{app:lower_variance}
Assume all variables $X_1,\dots,X_n$ have finite second moments. Since $\mathrm{Pa}(Y) \subseteq \mathrm{MB}(Y)$, apply the law of total variance with nested conditioning:

\begin{align*}
\mathrm{Var}\,\!\big(Y \mid \mathrm{Pa}(Y)\big)
&=
\mathbb{E}\,\!\Big[\mathrm{Var}\,\!\big(Y \mid \mathrm{MB}(Y)\big)\,\Big|\,\mathrm{Pa}(Y)\Big]
\;+\; 
\notag\\
&\quad\quad
\mathrm{Var}\,\!\Big(\mathbb{E}\,\!\big[Y \mid \mathrm{MB}(Y)\big]\;\Big|\;\mathrm{Pa}(Y)\Big)\\
&\;\ge\;
\mathbb{E}\,\!\Big[\mathrm{Var}\,\!\big(Y \mid \mathrm{MB}(Y)\big)\,\Big|\,\mathrm{Pa}(Y)\Big]
\end{align*}
Taking expectation over $\mathrm{Pa}(Y)$ yields
\[
\mathbb{E}_{\mathrm{Pa}(Y)}\!\Big[\,\mathrm{Var}\,\!\big(Y \mid \mathrm{Pa}(Y)\big)\,\Big]
\;\ge\;
\mathbb{E}_{\mathrm{Pa}(Y)}\!\Big[\,\mathbb{E}\,\!\big[\mathrm{Var}(Y \mid \mathrm{MB}(Y)) \,\big|\, \mathrm{Pa}(Y)\big]\,\Big]
\]
Hence,
\[
\mathbb{E}\,[\mathrm{Var}(Y \mid \mathrm{Pa}(Y))] \ge \mathbb{E}\,[\mathrm{Var}(Y \mid \mathrm{MB}(Y))]
\]
Hence, conditioning on the larger information set $\mathrm{MB}(Y)$ (parents, children, and co-parents) cannot increase the residual uncertainty about $Y$, and under standard regularity conditions, this is stated pointwise (a.s.) as
\[
\mathrm{Var}\,\!\big(Y \mid \mathrm{MB}(Y)\big)
\;\le\;
\mathrm{Var}\,\!\big(Y \mid \mathrm{Pa}(Y)\big).
\]
\section{Causal Graphs Extracted by LSTM-LiNGAM}
\label{app:lingam}

We report the Markov blankets and causal structures discovered by LSTM-LiNGAM for each building-floor dataset.  

\begin{table}[H]
\centering
\caption{Causal structures discovered by LiNGAM. 
Abbreviations: IT = indoor\_temperature, IH = indoor\_humidity, IL = indoor\_light\_intensity, 
IC = indoor\_co2, IP = indoor\_pressure, IV = indoor\_tvoc, 
OT = outdoor\_temperature, W = is\_work.}
\label{tab:lingam_graphs}

\begin{tabularx}{0.95\textwidth}{|c|X|X|X|X|}
\hline
Dataset & Parents & Children & Spouses & Markov Blanket \\
\hline
North L-A & -- & \{IC, OT\} & \{W, IP, IL, IH\} & \{IL, IC, IH, W, IP, OT\} \\
\hline
North L-B & \{IT, IL, IC, IP, OT, W\} & -- & -- & \{IT, IL, IC, OT, W, IP\} \\
\hline
Central L-A  & -- & \{OT\} & \{IP, IL\} & \{IP, OT, IL\} \\
\hline
Central L-B & \{IV, IP, W\} & \{IT, IH, IC, OT\} & \{W, IV, OT, IP, IT, IL\} & \{W, IV, IC, OT, IP, IL, IT, IH\} \\
\hline
South L-A & \{IT, IL, IP\} & \{IH, IV\} & \{IV, IL, W, IP, IC, IT\} & \{IH, IL, W, IT, IP, IC, IV\} \\
\hline
South L-B & \{IL, IP\} & \{IH, OT\} & \{IV, IT, IP, OT, IL\} & \{IV, IT, IH, IP, OT, IL\} \\
\hline
\end{tabularx}
\end{table}

\section{Detail of indy loss ablation study}
\label{app:indy}
\begin{table}[h]
  \caption{Performance of \sysname{}-NoIndy vs. original \sysname{} (NMSE).}
  \centering
  \small
  \begin{tabular}{lccccccccc}
    \toprule
    \multirow{2}{*}{Area} & \multirow{2}{*}{L} 
    & \multicolumn{2}{c}{32} & \multicolumn{2}{c}{16} & \multicolumn{2}{c}{8} & \multicolumn{2}{c}{Average} \\
    \cmidrule(lr){3-4}\cmidrule(lr){5-6}\cmidrule(lr){7-8}\cmidrule(lr){9-10}
    & & no-indy & indy & no-indy & indy & no-indy & indy & no-indy & indy \\
    \midrule
    \multirow{2}{*}{North} 
      & L-A & 0.145 & 0.081 & 0.238 & 0.129 & 0.205 & 0.143 & \multirow{2}{*}{0.207} & \multirow{2}{*}{0.192} \\
      & L-B & 0.149 & 0.142 & 0.162 & 0.162 & 0.161 & 0.159 &       &       \\
    \midrule
    \multirow{2}{*}{Central} 
      & L-A  & 0.135 & 0.138 & 0.160 & 0.147 & 0.161 & 0.135 & \multirow{2}{*}{0.232} & \multirow{2}{*}{0.210} \\
      & L-B & 0.263 & 0.238 & 0.278 & 0.263 & 0.323 & 0.307 &       &       \\
    \midrule
    \multirow{2}{*}{South} 
      & L-A & 0.299 & 0.285 & 0.279 & 0.279 & 0.280 & 0.261 & \multirow{2}{*}{0.237} & \multirow{2}{*}{0.214} \\
      & L-B & 0.251 & 0.270 & 0.277 & 0.282 & 0.289 & 0.278 &       &       \\
    \bottomrule
  \end{tabular}
  \label{tab:temp}
\end{table}

\section{Derivation of the Debiased Jacobian}
\label{app:debias-deriv}
We debias the contribution matrix to remove the influence of weights $g$. The encoder applies a global, sample-agnostic gate $g=\mathrm{softmax}(\alpha)\in(0,1)^p$ (Eq.~\ref{eq:gate}) and forms $\tilde X_t=g\odot X_t$. By the chain rule,
\begin{equation}
\label{eq:roc}
\frac{\partial Z}{\partial X_t}
\;=\;
\frac{\partial Z}{\partial \tilde X_t}\,\frac{\partial \tilde X_t}{\partial X_t}
\;=\;
\frac{\partial Z}{\partial \tilde X_t}\,\mathrm{diag}(g).
\end{equation}
Let $A_t := \bigl|\tfrac{\partial Z}{\partial \tilde X_t}\bigr|$. Since $g_j\ge 0$ and $\mathrm{diag}(g)$ rescales columns, we have the columnwise identity
\[
\bigl|A_t\,\mathrm{diag}(g)\bigr|
\;=\;
A_t\,\mathrm{diag}(g),
\]
hence
\[
J
\;=\;
\frac{1}{T}\sum_{t=1}^{T} \bigl|{\textstyle \frac{\partial Z}{\partial X_t}}\bigr|
\;=\;
\frac{1}{T}\sum_{t=1}^{T} A_t\,\mathrm{diag}(g)
\;=\;
\underbrace{\Bigl(\frac{1}{T}\sum_{t=1}^{T} A_t\Bigr)}_{\displaystyle \frac{1}{T}\sum_{t=1}^{T}\bigl|\frac{\partial Z}{\partial \tilde X_t}\bigr|}
\;\mathrm{diag}(g).
\]

Therefore, the \emph{debiased} Jacobian that removes the structural rescaling by $g$ is obtained by right-multiplication with $\mathrm{diag}(g)^{-1}$:
\[
\boxed{\quad
J_{\text{debiased}}
\;=\;
J\,\mathrm{diag}(g)^{-1}
\;=\;
\frac{1}{T}\sum_{t=1}^{T}\Bigl|\tfrac{\partial Z}{\partial \tilde X_t}\Bigr|
\quad}
\]

\begin{remark}
For clarity, we assume $h_\phi$ to be linear when deriving Eq.~\eqref{eq:roc}. 
In reality, since the network contains non-linear activation functions, the mapping is only piecewise linear. Consequently, the debiasing may not be perfectly accurate in a global sense, but it still provides a reasonable correction that mitigates the bias to a large extent.
\end{remark}

\section{Weighted derivatives of causal representation}
\label{app:contrib}
\begin{table}[ht]
\centering
\caption{Jacobian-based influence of raw features on each representation dimension (Original vs. Debiased)}
\label{tab:contrib}
\small
\begin{tabular}{lcccccccc}
\toprule
\multicolumn{9}{c}{Original Jacobian} \\
\midrule
$Z$ & \texttt{outdoor\_temperature} & \texttt{indoor\_light} & \texttt{indoor\_pressure} & \texttt{is\_work} & \texttt{indoor\_temperature} & \texttt{indoor\_co2} & \texttt{indoor\_humidity} & \texttt{indoor\_tvoc} \\
\midrule
$Z_0$ & 0.03127 & 0.00991 & 0.00802 & 0.00748 & 0.00736 & 0.00688 & 0.00473 & 0.00270 \\
$Z_1$ & 0.00496 & 0.00251 & 0.00472 & 0.00347 & 0.00195 & 0.00279 & 0.00065 & 0.00312 \\
$Z_2$ & 0.00147 & 0.00063 & 0.00242 & 0.00064 & 0.00100 & 0.00045 & 0.00048 & 0.00087 \\
$Z_3$ & 0.02558 & 0.01579 & 0.00568 & 0.01150 & 0.00968 & 0.00681 & 0.00348 & 0.00397 \\
$Z_4$ & 0.00147 & 0.00041 & 0.00063 & 0.00042 & 0.00052 & 0.00040 & 0.00249 & 0.00051 \\
\midrule
\multicolumn{9}{c}{Debiased Jacobian} \\
\midrule
$Z_0$ & 0.22649 & 0.07026 & 0.06907 & 0.05490 & 0.06183 & 0.05714 & 0.04049 & 0.02416 \\
$Z_1$ & 0.03596 & 0.01777 & 0.04068 & 0.02544 & 0.01638 & 0.02316 & 0.00554 & 0.02789 \\
$Z_2$ & 0.01062 & 0.00448 & 0.02083 & 0.00473 & 0.00837 & 0.00373 & 0.00408 & 0.00778 \\
$Z_3$ & 0.18528 & 0.11191 & 0.04891 & 0.08439 & 0.08127 & 0.05659 & 0.02972 & 0.03542 \\
$Z_4$ & 0.01063 & 0.00288 & 0.00544 & 0.00310 & 0.00436 & 0.00328 & 0.02129 & 0.00455 \\
\bottomrule
\end{tabular}
\end{table}

\end{document}